\documentclass[conference]{IEEEconf}
\IEEEoverridecommandlockouts
\usepackage{cite}
\usepackage{amsmath,amssymb,amsfonts}
\usepackage{algorithmic}
\usepackage{graphicx}
\usepackage{textcomp}
\usepackage{xcolor}
\usepackage{caption}
\usepackage{booktabs}

\usepackage{subcaption}
\graphicspath{ {./figures/} }

\def\BibTeX{{\rm B\kern-.05em{\sc i\kern-.025em b}\kern-.08em
    T\kern-.1667em\lower.7ex\hbox{E}\kern-.125emX}}
\begin{document}

\title{\textit{Multi S-graphs}: A Collaborative Semantic SLAM architecture.}

\author{
    Miguel Fernandez-Cortizas$^{1,2}$, Hriday Bavle$^{1}$, Jose Luis Sanchez-Lopez$^{1}$,\\ Pascual Campoy$^{2}$ and Holger Voos$^{1}$ 
    \thanks{$^{1}$Authors are with the Automation and Robotics Research Group, Interdisciplinary Centre for Security, Reliability, and Trust (SnT), University of Luxembourg, Luxembourg. Holger Voos is also associated with the Faculty of Science, Technology, and Medicine, University of Luxembourg, Luxembourg. \tt{\small{\{hriday.bavle, joseluis.sanchezlopez, holger.voos\}}@uni.lu}}
    \thanks{$^{2}$Authors are with the Computer Vision and Aerial Robotics Group (CVAR), Centre for Automation and Robotics (CAR), Escuela Tecnica Superior de Ingenieros Industriales (ETSII), Universidad Politécnica de Madrid, Spain. \tt\small{\{miguel.fernandez.cortizas, pascual.campoy\}@upm.es}}
    \thanks{*This work was partially funded by the Fonds National de la Recherche of Luxembourg (FNR), under the projects C19/IS/13713801/5G-Sky, by the European Union’s Horizon 2020 Project No. 101017258 SESAME, by European Union's Horizon Europe Project No. 101070254 CORESENSE, as well as project COPILOT ref. 2020/EMT6368, funded by the Madrid Government under the R\&D Synergic Projects Program and project INSERTION ref. ID2021-127648OBC32,  funded by the Spanish Ministry of Science and Innovation.}
    \thanks{For the purpose of Open Access, the author has applied a CC BY public copyright license to any Author Accepted Manuscript version arising from this submission.}
}
\maketitle
\begin{abstract}

Collaborative Simultaneous Localization and Mapping (CSLAM) is a critical capability for enabling multiple robots to operate in complex environments. Most CSLAM techniques rely on the transmission of low-level features for visual and LiDAR-based approaches, which are used for pose graph optimization. However, these low-level features can lead to incorrect loop closures, negatively impacting map generation. Recent approaches have proposed the use of high-level semantic information in the form of Hierarchical Semantic Graphs to improve the loop closure procedures and overall precision of SLAM algorithms. In this work, we present \textit{Multi S-Graphs}, an \textit{S-graphs}\cite{Bavle2022} based distributed CSLAM algorithm that utilizes high-level semantic information for cooperative map generation while minimizing the amount of information exchanged between robots. 
Experimental results demonstrate the promising performance of the proposed algorithm in map generation tasks.

\end{abstract}

\section{Introduction}

Collaborative Simultaneous Localization and Mapping (CSLAM) is a fundamental capability that enables multiple robots to operate in complex environments with multiple robots coordinately.

Most CSLAM techniques, such as \cite{Lajoie2020}\cite{zhong2022dcl}\cite{Huang2022}  are heavily based on the transmission of low-level features, such as keyframe descriptors, for both visual and LiDAR-based approaches. These low-level features constitute the core of the majority of the Pose Graph Optimization (PGO) SLAM based methods and relies on these low-level features for the creation and optimization of each Pose Graph. Using this low-level feature to align and extend the pose graphs created for each robot usually leads to incorrect loop closures; some works like \cite{Lajoie2020}  or \cite{mangelson2018pairwise} are focused on robustifying their loop closure algorithms to avoid incorrect loop closures that could ruin the overall map generation. The main problem about these methods emerges from the fact that the system has no awareness about what each low-level feature means, or if it has sense to create a loop closure between nodes or not. 

Lately, some SLAMs approaches like \textit{Hydra} \cite{Hughes2022} or \textit{S-Graphs+} \cite{Bavle2022} tend to deal with this issue of lack of awareness in the field of SLAM, betting for the use of Hierarchical Semantic Graphs during the generation of the Pose Graphs, to include high-level semantic information about the architectural components (Walls, Rooms, floors) into their ``mental model'', which can later be used to improve the loop closure procedures and to improve the overall precision of the SLAM algorithms.

However, as far as we know, these high-level semantic representations have not been used to improve the performance of multi-robot SLAM algorithms that can take advantage of this semantic knowledge to reduce the amount of information that has to be transmitted between agents and to robustify loop closures, pursuing the best overall mapping and localization quality.

In this work, we present \textit{Multi S-Graphs}, a LiDAR based distributed CSLAM algorithm that relies on high-level semantic information to generate a complete map of a building cooperatively exchanging a minimum amount of information between them.

The main contributions presented in this work are as follows:
\begin{enumerate}
    \item A novel distributed multi-robot SLAM architecture that relies on high-level semantic features for communicating information between agents.
    \item A hybrid descriptor that combines the fine-grained information of a pointcloud with semantic knowledge.
    \item A real-time CSLAM algorithm robust to multiple robot initialization, considering the multiple kidnapped robot problem.
    
\end{enumerate}

\begin{figure*}[!htb]
    \centering
    \includegraphics[width=0.80\textwidth]{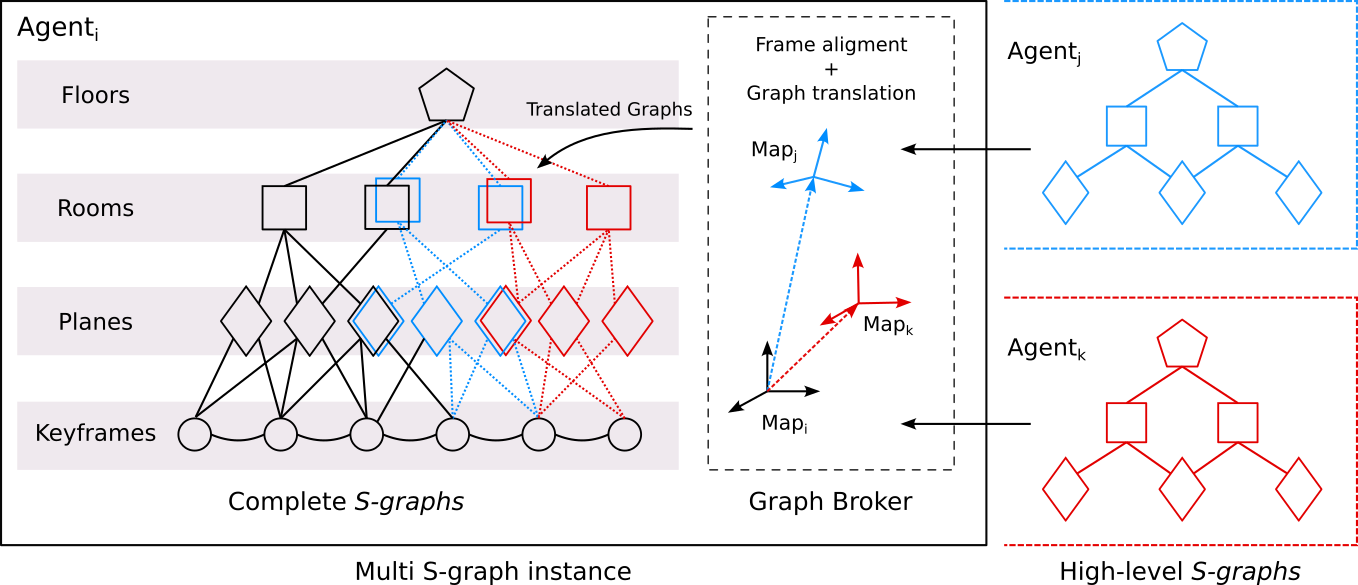}
    \caption{\textit{Multi S-graph} architecture schema viewed from $Agent_{i}$ perspective.  }
    \label{fig:multi_sgraph_schema}
    \vspace{-0.4cm}
\end{figure*}

\section{Related Work}

Although, the algorithm presented is a LiDAR based multi-robot SLAM pipeline, we will include Visual Based algorithms to further understand how the multi-robot approaches are accomplished within the field.

Currently, the vast majority of the multi-robot SLAM methods relies on Pose Graph Optimization (PGO) approaches in which the agents exchange information of the same type that each graph uses for generating the internal loop closures. 

In LiDAR-based approaches, Zhong et al. \cite{zhong2022dcl} and Huang et al.\cite{Huang2022} proposed a framework based on Scan Context Descriptors \cite{kim2018scan}. In \cite{zhong2022dcl} also presents a P2P communication protocol for exchanging the descriptors of each keyframe and uses Binarized Scan Contexts. In both works, each Robot runs its own PGO pipeline.



Within visual-based approaches, Deustch et al. \cite{Deutsch2016} proposed a framework that relies on a BoW of the keyframes obtained with an RGB-d camera. Lajoie et al. \cite{Lajoie2020} \cite{lajoie2023swarmslam} proposed a distributed CSLAM system based on NetVLAD descriptors. 
KIMERA multi \cite{tian23arxiv_kimeramultiexperiments}, also uses BoW and needs a \textit{Robust Distributed Initialization} to initialize all robot poses in a shared (global) coordinate frame.

Finally, Bernreiter et al. \cite{bernreiter2022framework} presented a centralized CSLAM method based on spectral graph waves, which consists of analyzing the SE(3) Pose graph of each robot and trying to find coincidences and discrepancies in the graph structure of each robot compared to the global graph. This algorithm does not rely on a specific sensor, but just on the pose graph generated.
 


\section{Colaborative S-Graphs}

\subsection{Nomenclature}

In this work, we present a distributed approach for multi-robot semantic SLAM. In our approach, we consider each robot (agent) that interacts in a 1 to N fashion. This means that each robot will interact with as many robots as possible independently, each robot will be denoted as Agent $A_i$. An schema of the architecture is shown in Fig. \ref{fig:multi_sgraph_schema}.

Each agent will run its own \textit{S-graphs} pipeline. \textit{S-Graphs} are four-layered optimizable hierarchical graphs
built online using 3D LiDAR measurements. The full details of the \textit{S-Graphs} we use in this work can be found in \cite{Bavle2022}. In brief, their four layers can be summarized as follows: 

\begin{itemize}
\item Keyframes Layer. It consists of robot poses factored as SE(3) nodes in the agent map frame $A_i$ with pairwise odometry measurements constraining them.
\item Walls Layer. It consists of the planar wall surfaces extracted from the 3D LiDAR measurements and factored
using minimal plane parameterization. The planes observed by their respective keyframes are factored using pose-plane constraints.
\item Rooms Layer: It consists of two-wall rooms or four-wall rooms, each constraining either two or four detected wall surfaces, respectively.
\item Floors Layer: It consists of a floor node 2 positioned in the center of the current floor.
\end{itemize}

From \textit{S-graphs}, we will only consider the following vertices: Rooms $R_{i,k}$, Planes $P_{i,k}$ and Keyframes $K_{i,k}$, where the $i$ index denoted the agent that contains this vertex in its own graph, and $k$ the index of the vertex.

Each vertex can be translated into different agents coordinated frames. We denote $^{A_j}V_{i,k}$ as the $k$-vertex $V$ of the robot $i$ expressed in the agent $j$ reference frame.

\subsection{Room descriptors}
In order to avoid errors aligning the robot positions in very symmetric situations, like a corridor with multiple rooms, one on side of the order, we cannot only rely on the structural information stored in the top layers of the \textit{S-graphs}, lower level information may be needed to break the symmetry and decide if two rooms are the same or not.

Compared to other LiDAR-based SLAM methods, \textit{S-graphs} does not take continuous snapshots of the pointcloud measures, these measures are very sparse, so using classical pointcloud feature-based pointcloud matching is not the most convenient way. In order to take advantage of the semantic information that each room contains, we decided to generate a hybrid descriptor that combines the fine-grained information of a pointcloud with high-level semantic knowledge, a  \textit{Room Descriptor}. 

For generating these descriptors, we use an Scan Context descriptor \cite{kim2018scan} approach, an egocentric, yaw-invariant descriptor. This descriptor has achieved satisfactory results in multiple LiDAR odometry, and SLAM works because of its simplicity and fast generation. However, one of the drawbacks of these descriptors is the sensitiveness of these descriptions to translation. 

Here, we take advantage of the semantic information in the room, by generating a scan context from the centre of each room, avoiding translation errors. To generate the Room Descriptor, we need a \textit{Room Keyframe}, which is built by combining all point clouds obtained by the robot from within a room. Each Room Keyframe $Rk_i$ can be expressed as:

\begin{equation}
    Rk_{i} = U \{^{R_{i}}K_j\} \quad ;\quad  \forall j \;|\; K_j \in R_{i}
\end{equation}

where $^{R_{i}}K_j$ represents the pointcloud associated with the keyframe $K_j$ in the $R_i$ frame (a frame located in the center of the room $i$).

\begin{figure}[!htb]
    \centering
    \includegraphics[width=0.48\textwidth]{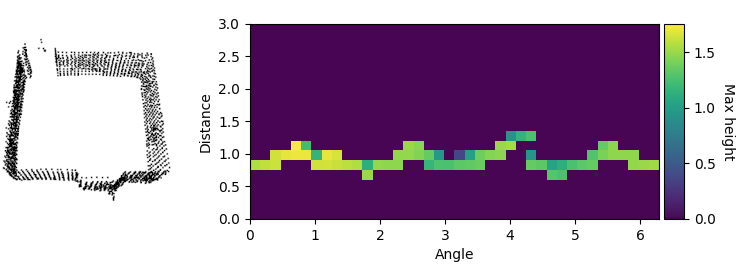}
    \caption{Room Descriptor (right) obtained from a Room Keyframe (left).}
    \label{fig:room_descriptor}
\end{figure}

To obtain the Room Descriptor $Rd_i$ from a Room Keyframe $Rk_i$ a downsample of the $Rk_i$ with a voxel size of 0.1 $m$ is done to homogenize the number of points that each keypoint has independently of the number of keyframes associated with each room. Finally, the scan context descriptor of each Room Keyframe $Rk_i$ is computed to create the Room Descriptor $Rd_i$:

\begin{equation}
    Rd_{i} = SC(\phi(Rk_i))
\end{equation}
where $\phi(Rk_i)$ represents the downsample of the keyframe, and $SC(\vec{X}):R^{n\times3} \rightarrow R^{n_s\times n_r}$ is the Scan Context obtention from a pointcloud.
An example of this room descriptor is shown in Fig. \ref{fig:room_descriptor}.

The use of this descriptor will make the difference in the alignment and further optimization steps.

\subsection{Robots alignment} 

As we start from the problem of multiple kidnapped robots, no initial estimation of the relative positions of the robots is provided. If we try to align the complete pointclouds obtained from the multiple robots, we will meet the global registration problem, which, combined with the noise of each pointcloud and no prior information of a possible transformation, leads to unsuitable alignments. 

In order to generate good candidates for alignment, we leverage in the Room Descriptors to generate a global alignment of each robot coordinated system, which is crucial for the further graph sharing and collective optimization.

The module in charge of finding this relative transformation between the robots is \textit{Graph Broker}.

 In order to compute this transformation, we perform a two-step process:
\begin{enumerate}
    \item Descriptor matching: The broker receives and stores the room descriptors of the rest of the agents, trying to find a suitable match. 
    \item Fine alignment: Whenever a match is found between robot and other agents' keyframes, it tries to obtain an improved transform from the room keyframe using a VGICP\cite{vgicp} registration algorithm. The validity of the relative transform is determined by alignment distance and matching threshold $d_t$. If suitable, the rest of the graph information can be transformed into the local robot frame for optimization.
    


\end{enumerate}

\subsection{Multi-robot mapping} 

Whenever a transformation between robots is found, then the top layers of the \textit{S-graphs} can be shared and incorporated into the other robot graph.

In this approach, there are 2 types of graph vertices that are exchanged:
\begin{itemize}
    \item Room vertices: Each room vertex includes the $SE(3)$ transformation of the Room center in the agent frame $^{A_i}R_{i,k}$ .
    \item Plane vertices: Each plane vertex includes the $\vec{n}$ normal to the plane and the distance $d$ from this plane to the agent frame $A_i$
\end{itemize}

These vertices are joined with edges that relate the planes that conform each room.

\begin{figure*}[!htb]
     \centering
    \includegraphics[width=0.67\textwidth]{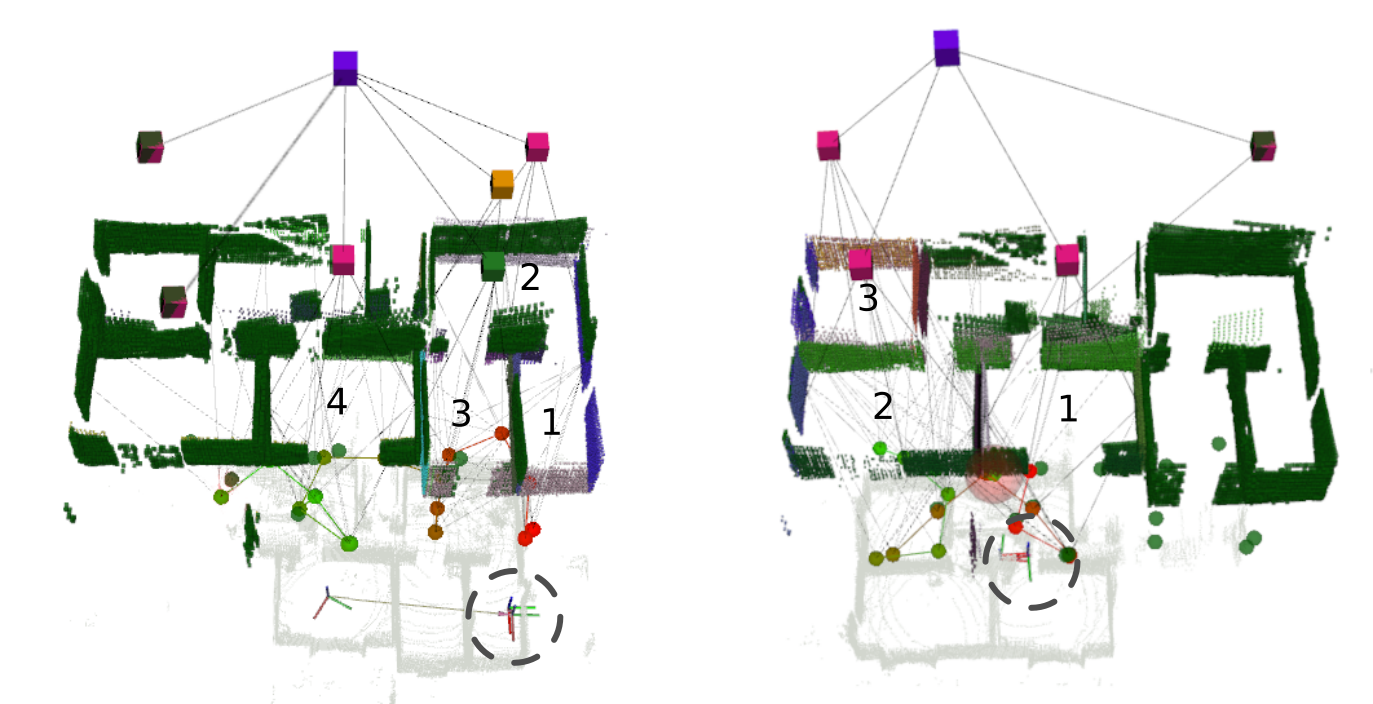}
        \caption{Collaborative maps created for two robots, displaying green Rooms and Planes provided by the other agent. Dotted circles indicate the initial position of each robot, while room numbers indicate their navigation order.}
        \label{fig:experiments}
        \vspace{-0.4cm}
\end{figure*}

The optimization pipeline consists of 3 steps that repeat: 

\subsubsection{Vertex transform}
After the transformation between agents $^{A_i}T_{A_j}$ is found, the vertices that came from the $j$ agent can be transformed and added to the graph of the agent $i$.

The rooms transforms are:
\begin{equation}
    ^{A_i}R_{j,k} = ^{A_i}T_{A_j} \; ^{A_j}R_{j,k} \quad ; T_{j,k} \in SE(3)
\end{equation}

Considering each plane as follows:
\begin{equation}
^{A_i}P_{i,k} = \begin{bmatrix}
      ^{A_i}\mathbf{n}_k\\ ^{A_i} d_k
\end{bmatrix}
\end{equation}
where $^{A_i}\mathbf{n_k}$  is the normal vector to the plane in the $i$-agent map frame, and $d_k$ is the distance between this plane and the $i$-agent origin of coordinates.

The plane transforms are:
\begin{equation}
^{A_i}P_{i,k} = \begin{bmatrix}
      ^{A_i}\mathbf{n}_k\\ ^{A_i} d_k
\end{bmatrix} = \begin{bmatrix}
          ^{A_i}\mathbf{R}_{A_j} & 0 \\ -^{A_j}\mathbf{t}_{A_j} & 1
      \end{bmatrix} \begin{bmatrix}
      ^{A_j}\mathbf{n}_k\\ ^{A_j} d_k
\end{bmatrix}
\end{equation}

\subsubsection{Data association}
Whenever the external vertices are transformed into the corresponding agent frame, a data association process is performed. In this step, similarities between vertices are searched for, no matter if they are internal or external vertices. If two vertices are similar, then an association is made and a new factor is created between them. Further details on data association criteria can be found in \cite{Bavle2022}.

\subsubsection{Graph Optimization}
After this data association, the rest of the optimization process is similar to the one used in \textit{S-graphs} \cite{Bavle2022}.

\section{Experimental Results}
In our experiments, we generate a map of a building floor collaboratively with two robots.
Each robot starts at a different place and is unknown to the rest.

During the experiment, we divided a floor into two parts to be explored; the first robot covers the right-hand rooms of the floor and the second one covers the left-hand rooms. A central room is covered for both robots to have a common room, which could lead to the alignment of the robot frames. The data of the experiment were collected using a Boston Dynamics Spot carrying a Velodyne VLP-{16} in a real construction site.

As is shown in Fig. \ref{fig:experiments} both robots are capable of integrating the information collected by the other robot into its own graph, and both robots optimize its own graph by taking into account the information provided by the counterpart. Table \ref{tab:times_table} compares the mapping times between the \textit{S-graphs} with one robot and our proposal with two robots, to map one area.

\begin{table}[htb]
    \centering
    \begin{tabular}{cccc}
         \toprule
         Experiment & \textit{S-graphs+}& \textit{Multi S-graphs} & Overlapping time.\\
         \toprule
         Construction 1 & 203 s & 123 s & 22s (18\%) \\
        \bottomrule
    \end{tabular}
    \caption{Elapsed time in generating a complete map of an area, including overlapping time for matching when both robots explore the same rooms.}
    \label{tab:times_table}
\end{table}
\vspace{-0.3cm}
\section{Conclussions and Future Work}

In this work a distributed multi-robot SLAM  algorithm is presented, leveraging in the semantic features extracted by the \textit{S-graphs} SLAM algorithm, in order to filter and reduce the amount of data that has to be transmitted between robots. This algorithm considers the kidnapped robot problem for all their robots, and is able to align the maps of the different robots taking advantage of the Room Keyframe descriptor, which combines semantic information with low-level features.
We have tested this algorithm for a map generation task, achieving promising results. 

In this work, each robot optimizes its own graph with the information obtained by the others, but the optimization that each one mades is not feedbacked to the rest of the agents. In order to achieve the best results, this optimization should be transmitted to the rest in order to achieve a global graph optimization. Moreover, a thorough experimental evaluation in different simulated and real environments has to be done for measuring the performance of the proposed algorithm.

\bibliographystyle{IEEEtran}
\bibliography{bibliography}

\end{document}